\documentclass[runningheads]{llncs}
\usepackage{graphicx}
\usepackage{amsmath,amssymb} 
\usepackage{diagbox}
\usepackage{hyperref}
\usepackage{algorithm2e}

\usepackage{color}
\usepackage{float}
\usepackage[width=122mm,left=12mm,paperwidth=146mm,height=193mm,top=12mm,paperheight=217mm]{geometry}

\begin{document}
\pagestyle{headings}
\mainmatter

\title{UnrealCV: Connecting Computer Vision to Unreal Engine} 

\titlerunning{UnrealCV: Connecting Computer Vision to Unreal Engine}

\authorrunning{Weichao Qiu, Alan Yuille}

\author{Weichao Qiu, Alan Yuille}
\institute{
Johns Hopkins University, Baltimore, MD, USA\\
\email{ \{qiuwch, alan.l.yuille\}@gmail.com }
}

\maketitle
\vspace{-1em}
\begin{abstract}

Computer graphics can not only generate synthetic images and ground truth but it also offers the possibility of constructing {\it virtual worlds} in which: (i) an agent can perceive, navigate, and take actions guided by AI algorithms, (ii) properties of the worlds can be modified (e.g., material and reflectance), (iii) physical simulations can be performed, and (iv) algorithms can be learnt and evaluated. But creating realistic virtual worlds is not easy. The game industry, however, has spent a lot of effort creating 3D worlds, which a player can interact with. So researchers can build on these resources to create virtual worlds, provided we can access and modify the internal data structures of the games. To enable this we created an open-source plugin {\it UnrealCV}\footnote{Project website: http://unrealcv.github.io} for a popular game engine Unreal Engine 4 (UE4). We show two applications: (i) a proof of concept image dataset,  and (ii) linking Caffe with the virtual world to test deep network algorithms.

\end{abstract}

\vspace{-1em}
\section{Introduction}

Computer vision has benefitted enormously from large datasets\cite{Deng:2009jn,Everingham:2009dq}. They enable the training and testing of complex models such as deep networks\cite{Krizhevsky:2012wl}. But performing annotation is costly and time consuming so it is attractive to make synthetic datasets which contain large amounts of images and detailed annotation. These datasets are created by modifying open-source movies\cite{Butler:2012cv} or by constructing a 3D world\cite{Ros:2016wf,Gaidon:2016vn}. Researchers have shown that training on synthetic images is helpful for real world tasks\cite{Su:2015tm,Peng:2014tg,Hattori:2015hw,marin2010learning,vazquez2014virtual}. Robotics researchers have gone further by constructing 3D worlds for robotics simulation, but they emphasize physical accuracy rather than visual realism. This motivates the design of realistic {\it virtual worlds} for computer vision where an agent can take actions guided by AI algorithms, properties of the worlds can be modified, physical simulations can be performed, and algorithms can be trained and tested. Virtual worlds have been used for autonomous driving\cite{Chen:2015uu}, naive physics simulations\cite{Battaglia:2013fm} and evaluating surveillance system\cite{taylor2007ovvv}. But creating realistic virtual worlds is time consuming.

\begin{figure}
    \centering
    \includegraphics[width=\textwidth]{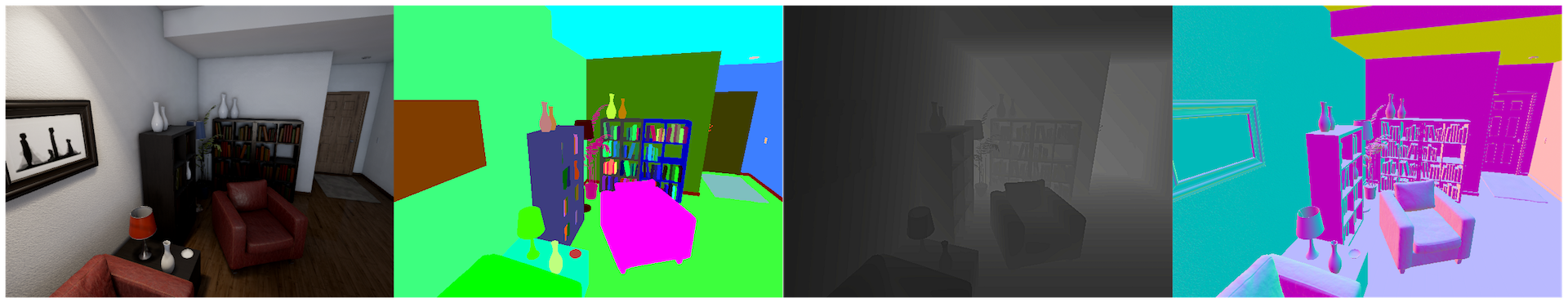}
    \caption{A synthetic image and its ground truth generated using UnrealCV. The virtual room is from technical demo RealisticRendering, built by Epic Games. From left to right are the synthetic image, object instance mask, depth, surface normal \label{fig:teaser}}
\end{figure}

The video game industry has developed many tools for constructing 3D worlds, such as libraries of 3D object models. These 3D worlds are already realistic and the popularity of games and Virtual Reality (VR) drives towards even greater realism. So modifying games and movies is an attractive way to make virtual worlds\cite{Chen:2015uu}. But modifying individual games is time-consuming and almost impossible for proprietary games. Hence our strategy is to modify a game engine, so that all the games built on top of it can be used. We develop a tool, UnrealCV, which can be used in combination with a leading game engine, Unreal Engine 4 (UE4), to use the rich resources in the game industry. UnrealCV can also be applied to 3D worlds created for virtual reality, architecture visualization, and computer graphics movies, provided they have been created using UE4. More precisely, UnrealCV provides an UE4 plugin. If a game, or any 3D world, is compiled with this plugin then we can create a virtual world where we can access and modify the internal data structures. This allows us to connect AI programs, like Caffe, to it and use a set of commands provided by UnrealCV to obtain groundtruth, control an agent, and so on. Fig.~\ref{fig:teaser} shows a synthetic image and its ground truth generated using UnrealCV.

We stress that we provide an open-source tool to help create new virtual worlds, which differs from work which produces a single virtual world\cite{Chen:2015uu} or creates synthetic datasets\cite{Ros:2016wf,Gaidon:2016vn}. We hope that our work can help build a bridge between Unreal Engine and computer vision researchers.

\vspace{-1em}

\vspace{-1em}
\section{Related Work \label{sec:related_work}}

Virtual worlds have been widely used in robotics research and many robotics simulators have been built\cite{Koenig:2004dh,Todorov:2012kg}. But these focus more on physical accuracy than visual realism, which makes them less suitable for computer vision researchers. Unreal Engine 2 (UE2) was used for robotics simulation in USARSim\cite{Carpin:2007ff}, but UE2 is no longer available and USARSim is no longer actively maintained.

Computer vision researchers have created large 3D repositories and virtual scenes\cite{Chang:2015uq,Handa:2015up,Mottaghi:2016tj,Choi:2016wf}. Note that these 3D resources can be used in the combined Unreal Engine and UnrealCV system.


Games and movies have already been used in computer vision research. An optical flow dataset was generated from the open source movie Sintel\cite{Butler:2012cv}. TORCS, an open source racing game, was converted into a virtual world and used to train an autonomous driving system\cite{Chen:2015uu}. City scenes were built\cite{Ros:2016wf,Gaidon:2016vn} using the Unity game engine to produce synthetic images. By contrast, UnrealCV extends the functions of Unreal Engine and provides a tool for creating virtual worlds instead of generating a synthetic image/video dataset or producing a single virtual world.

\section{Unreal Engine \label{sec:unreal_engine}}

A game engine contains the components shared by many video games, such as rendering code and design tools. Games built using a game engine combine components from the engine with the game logic and 3D models. So modifying a game engine can affect all games built on top of it.

\begin{figure}[h]
    \includegraphics[width=\textwidth]{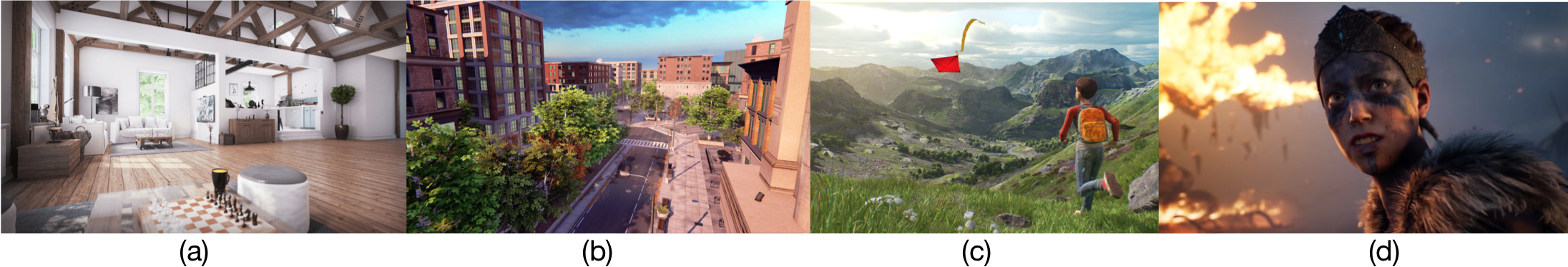}
    \caption{Images produced by UE4, (a)(b) An architectural visualization and an urban city scene from Unreal Engine marketplace. (c) An open-source outdoor scene KiteRunner. (d) A digital human from the game Hellblade, shown in the conference GDC2016
    \label{fig:unreal_images}}
\end{figure}

We chose UE4 as our platform for these reasons: (I) It is fully open-source and can be easily modified for research. (II) It has the ability to produce realistic images, see Fig.~\ref{fig:unreal_images}. (III) It provides nice tools and documentation for creating a virtual world. These tools integrate well with other commercial software and well maintained. (IV) It has a broad impact beyond the game industry and is a popular choice for VR and architectural visualization, so high-quality 3D contents are easily accessible.

\section{UnrealCV \label{sec:unrealcv}}


UE4 was designed to create video games. To use it to create virtual worlds, a few modifications are required: (I) The camera should be programmably controlled, instead of by the keyboard and mouse, so that an agent can explore the world. (II) The internal data structure of the game needs to be accessed in order to generate ground truth. (III) We should be able to modify the world properties, such as lighting and material.

UnrealCV extends the function of UE4 to help create virtual worlds. More specifically, UnrealCV achieves this goal by a plugin for UE4. Compiling a game with the plugin installed embeds computer vision related functions to produce a virtual world. Any external program can communicate with this virtual world and use a set of commands provided by UnrealCV to perform various tasks. For example, the command \verb|vget /camera/0/rotation| can retrieve the rotation of the first camera in the scene.

{\noindent \bf Architecture}

\begin{figure}[h]
	\begin{center}
    \includegraphics[width=0.7\linewidth]{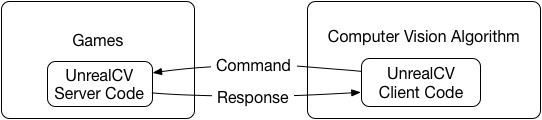}
    \end{center}
    \caption{The UnrealCV server is an UE4 plugin embedded into a game during compilation. An external program uses the UnrealCV client to communicate with the game.}
    \label{fig:pipeline}
\end{figure}

UnrealCV consists of two parts. The first is the UnrealCV server, which is embedded into a virtual world to access its internal data structure. The second is the UnrealCV client whose function is provided by a library which can be integrated into any external program, like Caffe, enabling the program to send commands defined by UnrealCV to the server to perform various tasks. The architecture is shown in Fig.~\ref{fig:pipeline}.

\vspace{-1em}

The UnrealCV server is an UE4 plugin. After installing the plugin to UE4, the UnrealCV server code will be embedded into a game during compilation. The server will start when the game launches and wait for commands. The UnrealCV client uses a socket to communicate with the server. We implemented the client code for Python and MATLAB. Socket is a method of communicating between programs and is universal across programming languages and operating systems. So it is easy to implement a client for any language and platform that can support socket.

The server and client communicate using a plain text protocol. The client sends an UnrealCV command to the server and waits for a response. The command can be used to do various tasks. It can apply force to an object; can modify the world by changing the lighting or object position; can get images and annotation from the world. For example, the commands \verb|vget /camera/0/image| and \verb|vget /camera/0/depth| can get the image and depth ground truth. The command will save image as PNG file and return its filename. Depth will be saved as high dynamical range (HDR) image file, since the pixel value of PNG is limited within $ [0 \dots 255] $. The command \verb|vset /camera/0/position 0 0 0| sets the camera position to [0, 0, 0]. An UnrealCV command contains two parts. The first part is an action which can be either \verb|vget| or \verb|vset|. The \verb|vget| means getting information from the scene without changing anything and \verb|vset| means changing some property of the world. The second part is an URI (Uniform Resource Identifier) representing something that UnrealCV can control. The URI is designed in a hierarchical modular structure which can be easily extended.

{\noindent \bf Features}  The design of UnrealCV gives it three features:

{\it Extensiblity:} The commands are defined in a hierarchical modular way. Setting the light intensity can be achieved by \verb|vset /light/[name]/intensity| to change the light color, a new command \verb|vset /light/[name]/color| can be added without affecting the existing commands. UnrealCV is open-source and can be extended by us or other researchers.

{\it Ease of Use:} Since we provide compiled binaries of some virtual worlds, such as a realistic indoor room,  using UnrealCV is as simple as downloading a game and running it. Hence researchers can use UnrealCV without knowledge of UE4. The design supports cross-platform and multi-languages (Python, MATLAB). It is straightforward to integrate UnrealCV with external programs and we show an example with Caffe in Sec.~\ref{sec:examples}.

{\it Rich Resources:} UnrealCV only uses the standard Application Programming Interface (API) of UE4, making it compatible with games built with UE4. We will provide virtual worlds with UnrealCV integrated and also host a model zoo to share virtual worlds created by the community.

\vspace{-1em}
\section{Applications \label{sec:examples}}

In this section we created a virtual world based on the UE4 technical demo RealisticRendering\footnote{\scriptsize \url{https://docs.unrealengine.com/latest/INT/Resources/Showcases/RealisticRendering/}} which contains an indoor room with sofa, TV, table, bookshelves, floor lamp, etc. The virtual world can be downloaded from our project website. We demonstrate two applications of this virtual world in this section.

{\noindent \bf Generating a synthetic image dataset}

\begin{figure}[h]
	\includegraphics[width=\textwidth]{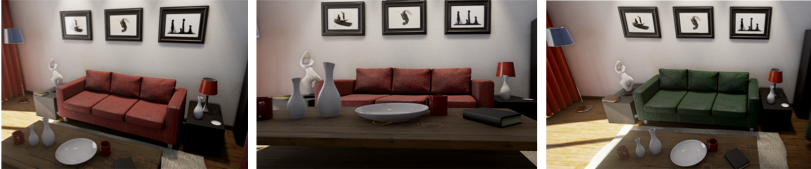}
	\caption{Images with different camera height and different sofa color.	\label{fig:rr_images}}
\end{figure}

We use a script to generate a synthetic image dataset from the virtual world. Images are taken using random camera positions. The camera is set to two different heights, human eye level and a Roomba robot level. The lighting, material property and object location can also be changed to increase the variety of the data, or to diagnose the strengths and weaknesses of an algorithm. Images with different camera height and sofa color can be seen in Fig.~\ref{fig:rr_images}. Ground truth, such as depth, surface normals and object instance masks, is generated together with the images, shown in Fig.~\ref{fig:teaser}. The ability to generate rich ground truth is particularly useful for training and testing algorithms which perform multiple tasks and for detailed understanding of a scene. The UnrealCV commands used to generate this synthetic image dataset are shown in Alg.~\ref{fig:commands}. The synthetic images are on our website and a tutorial shows how to generate them step-by-step. 

\RestyleAlgo{boxruled}
\begin{algorithm}[h]
	\scriptsize

	vget /objects \tcp*{Get objects information}

	\For{all camera position}{
	\tcc{Set the virtual camera position}
	vset /camera/0/location [x] [y] [z]\;
	vset /camera/0/rotation [yaw] [pitch] [roll]\;

	\tcc{Get image and ground truth}
	vget /camera/0/image\;
	vget /camera/0/depth, vget /camera/0/object\_mask\;
	}
\caption{Generate a synthetic image dataset from a virtual world \label{fig:commands}}
\end{algorithm}

{\noindent \bf Diagnosing a deep network algorithm}

We take a Faster-RCNN model\footnote{We use the implementation: \url{https://github.com/rbgirshick/py-faster-rcnn}} trained on PASCAL and test it in the virtual world by varying rendering configurations. The testing code uses the UnrealCV client to control the camera in the virtual world and the Faster-RCNN code tries to detect the sofa from different views. We moved the position of the camera but constrained it to always point towards the sofa shown in Fig.~\ref{fig:rr_images}. We got the object instance mask of the sofa and converted it into ground truth bounding box for evaluation. Human subject can easily detect the sofa from all the viewpoints. The Average Precision (AP) result shows surprisingly large variation as a function of viewpoint, see Tab.~\ref{tab:sofa_result}. For each az/el combination, the distance from the camera to the sofa was varied from 200cm to 290cm. The symbol ``-'' means the sofa is not visible from this viewpoint. More generally, we can vary parameters such as lighting, occlusion level, and camera viewpoint to thoroughly test an algorithm.

\setlength{\tabcolsep}{5pt}

\begin{table}[h]
\footnotesize
	\centering
	\begin{tabular}{|c|ccccc|}
	\hline
  \diagbox{Elevation}{Azimuth} &    90 &   135 &   180 &   225 &   270\\
\hline
0 &     - &     0.713 &     0.769 &     0.930 &     0.319\\
30 &     0.900 &     1.000 &     0.588 &     1.000 &     0.710\\
60 &     0.255 &     0.100 &     0.148 &     0.296 &     0.649\\	
\hline
	\end{tabular}
	\vspace{1em}
	\caption{The Average Precision (AP) when viewing the sofa from different viewpoints. Observe the AP varies from $0.1$ to $1.0$ showing the sensitivity to viewpoint. This is perhaps because the biases in the training cause Faster-RCNN to favor specific viewpoints.
	\label{tab:sofa_result}}
\end{table}

\vspace{-1em}
\section{Conclusion}

This paper has presented a tool called UnrealCV which can be plugged into the game engine UE4 to help construct realistic virtual worlds from the resources of the game, virtual reality, and architecture visualization industries. These virtual worlds allow us to access and modify the internal data structures enabling us to extract groundtruth, control an agent, and train and test algorithms. Using virtual worlds for computer vision still has  challenges, e.g., the variability of 3D content is limited, internal structure of 3D mesh is missing, realistic physics simulation is hard, and transfer from synthetic images remains an issue. But more realistic 3D contents will be available soon due to the advance of technology and the rising field of VR. As an industry leader, UE4 will benefit from this trend. UnrealCV is an open-source tool and we hope other researchers will use it and contribute to it.

{\noindent \textbf{Acknowledgment}} We would like to thank Yi Zhang, Austin Reiter, Vittal Premachandran, Lingxi Xie and Siyuan Qiao for discussion and feedback. This project is supported by the Intelligence Advanced Research Projects Activity (IARPA) with contract D16PC00007.

\bibliographystyle{splncs03}
\bibliography{unrealcv1}

\begin{thebibliography}{10}
\providecommand{\url}[1]{\texttt{#1}}
\providecommand{\urlprefix}{URL }

\bibitem{Battaglia:2013fm}
Battaglia, P.W., Hamrick, J.B., Tenenbaum, J.B.: Simulation as an engine of
  physical scene understanding. Proceedings of the National Academy of Sciences
   110(45),  18327--18332 (2013)

\bibitem{Butler:2012cv}
Butler, D.J., Wulff, J., Stanley, G.B., Black, M.J.: A naturalistic open source
  movie for optical flow evaluation. In: European Conference on Computer
  Vision. pp. 611--625. Springer (2012)

\bibitem{Carpin:2007ff}
Carpin, S., Lewis, M., Wang, J., Balakirsky, S., Scrapper, C.: Usarsim: a robot
  simulator for research and education. In: Proceedings 2007 IEEE International
  Conference on Robotics and Automation. pp. 1400--1405. IEEE (2007)

\bibitem{Chang:2015uq}
Chang, A.X., Funkhouser, T., Guibas, L., Hanrahan, P., Huang, Q., Li, Z.,
  Savarese, S., Savva, M., Song, S., Su, H., et~al.: Shapenet: An
  information-rich 3d model repository. arXiv preprint arXiv:1512.03012  (2015)

\bibitem{Chen:2015uu}
Chen, C., Seff, A., Kornhauser, A., Xiao, J.: Deepdriving: Learning affordance
  for direct perception in autonomous driving. In: Proceedings of the IEEE
  International Conference on Computer Vision. pp. 2722--2730 (2015)

\bibitem{Choi:2016wf}
Choi, S., Zhou, Q.Y., Miller, S., Koltun, V.: A large dataset of object scans.
  arXiv preprint arXiv:1602.02481  (2016)

\bibitem{Deng:2009jn}
Deng, J., Dong, W., Socher, R., Li, L.J., Li, K., Fei-Fei, L.: Imagenet: A
  large-scale hierarchical image database. In: Computer Vision and Pattern
  Recognition, 2009. CVPR 2009. IEEE Conference on. pp. 248--255. IEEE (2009)

\bibitem{Everingham:2009dq}
Everingham, M., Van~Gool, L., Williams, C.K., Winn, J., Zisserman, A.: The
  pascal visual object classes (voc) challenge. International journal of
  computer vision  88(2),  303--338 (2010)

\bibitem{Gaidon:2016vn}
Gaidon, A., Wang, Q., Cabon, Y., Vig, E.: Virtual worlds as proxy for
  multi-object tracking analysis. arXiv preprint arXiv:1605.06457  (2016)

\bibitem{Handa:2015up}
Handa, A., Patraucean, V., Badrinarayanan, V., Stent, S., Cipolla, R.:
  Scenenet: Understanding real world indoor scenes with synthetic data. arXiv
  preprint arXiv:1511.07041  (2015)

\bibitem{Hattori:2015hw}
Hattori, H., Naresh~Boddeti, V., Kitani, K.M., Kanade, T.: Learning
  scene-specific pedestrian detectors without real data. In: Proceedings of the
  IEEE Conference on Computer Vision and Pattern Recognition. pp. 3819--3827
  (2015)

\bibitem{Koenig:2004dh}
Koenig, N., Howard, A.: Design and use paradigms for gazebo, an open-source
  multi-robot simulator. In: Intelligent Robots and Systems, 2004.(IROS 2004).
  Proceedings. 2004 IEEE/RSJ International Conference on. vol.~3, pp.
  2149--2154. IEEE (2004)

\bibitem{Krizhevsky:2012wl}
Krizhevsky, A., Sutskever, I., Hinton, G.E.: Imagenet classification with deep
  convolutional neural networks. In: Advances in neural information processing
  systems. pp. 1097--1105 (2012)

\bibitem{marin2010learning}
Marin, J., V{\'a}zquez, D., Ger{\'o}nimo, D., L{\'o}pez, A.M.: Learning
  appearance in virtual scenarios for pedestrian detection. In: Computer Vision
  and Pattern Recognition (CVPR), 2010 IEEE Conference on. pp. 137--144. IEEE
  (2010)

\bibitem{Mottaghi:2016tj}
Mottaghi, R., Rastegari, M., Gupta, A., Farhadi, A.: " what happens if..."
  learning to predict the effect of forces in images. arXiv preprint
  arXiv:1603.05600  (2016)

\bibitem{Peng:2014tg}
Peng, X., Sun, B., Ali, K., Saenko, K.: Learning deep object detectors from 3d
  models. In: Proceedings of the IEEE International Conference on Computer
  Vision. pp. 1278--1286 (2015)

\bibitem{Ros:2016wf}
Ros, G., Sellart, L., Materzynska, J., Vazquez, D., Lopez, A.M.: The synthia
  dataset: A large collection of synthetic images for semantic segmentation of
  urban scenes. In: Proceedings of the IEEE Conference on Computer Vision and
  Pattern Recognition. pp. 3234--3243 (2016)

\bibitem{Su:2015tm}
Su, H., Qi, C.R., Li, Y., Guibas, L.J.: Render for cnn: Viewpoint estimation in
  images using cnns trained with rendered 3d model views. In: Proceedings of
  the IEEE International Conference on Computer Vision. pp. 2686--2694 (2015)

\bibitem{taylor2007ovvv}
Taylor, G.R., Chosak, A.J., Brewer, P.C.: Ovvv: Using virtual worlds to design
  and evaluate surveillance systems. In: 2007 IEEE Conference on Computer
  Vision and Pattern Recognition. pp. 1--8. IEEE (2007)

\bibitem{Todorov:2012kg}
Todorov, E., Erez, T., Tassa, Y.: Mujoco: A physics engine for model-based
  control. In: 2012 IEEE/RSJ International Conference on Intelligent Robots and
  Systems. pp. 5026--5033. IEEE (2012)

\bibitem{vazquez2014virtual}
Vazquez, D., Lopez, A.M., Marin, J., Ponsa, D., Geronimo, D.: Virtual and real
  world adaptation for pedestrian detection. IEEE transactions on pattern
  analysis and machine intelligence  36(4),  797--809 (2014)

\end{thebibliography}
\end{document}